\documentclass[a4paper, 10pt, conference]{ieeeconf}      
\usepackage{FG2025}

\FGfinalcopy 

\IEEEoverridecommandlockouts                              
\usepackage{amsmath,amsfonts,bm}

\def\eqref#1{equation~\ref{#1}}

\def\1{\bm{1}}

\DeclareMathAlphabet{\mathsfit}{\encodingdefault}{\sfdefault}{m}{sl}
\SetMathAlphabet{\mathsfit}{bold}{\encodingdefault}{\sfdefault}{bx}{n}

\usepackage[pdftex]{graphicx}
\usepackage{graphics} 
\usepackage{epsfig} 
\usepackage{times} 
\usepackage{amsmath} 
\usepackage{amssymb}  
\usepackage{lipsum}  
\usepackage{hyperref}

\usepackage{algorithm}
\usepackage{algpseudocode}

\usepackage{caption}

\usepackage{booktabs}
\usepackage{multicol}
\usepackage{multirow}

\usepackage{fancyhdr}
\usepackage{url}
\usepackage{verbatim}

\usepackage{wrapfig}

\usepackage{svg}
\usepackage{subcaption}

\usepackage{etoolbox}
\usepackage{pgf}
\usepackage{colortbl}

\def\FGPaperID{195} 

\title{\method: A Transformer for Multimodal, Multi-Party Social Signal Prediction  with Person-aware Blockwise Attention
}

\author{\parbox{16cm}{\centering
    {\large Yiming Tang, Abrar Anwar, and Jesse Thomason}\\
    {\normalsize
    University of Southern California}}
}

\newcommand{\method}{M3PT}
\newcommand{\methodlong}{\method\ (\textbf{M}ulti-\textbf{M}odal, \textbf{M}ulti-\textbf{P}arty \textbf{T}ransformer)}

\begin{document}

\ifFGfinal
\thispagestyle{empty}
\pagestyle{empty}
\else
\author{Anonymous FG2025 submission\\ Paper ID \FGPaperID \\}
\pagestyle{plain}
\fi
\maketitle

\begin{abstract}
Understanding social signals in multi-party conversations is important for human-robot interaction and artificial social intelligence. 
Social signals include body pose, head pose, speech, and context-specific activities like acquiring and taking bites of food when dining. 
Past work in multi-party interaction tends to build task-specific models for predicting social signals.
In this work, we address the challenge of predicting multimodal social signals in multi-party settings in a single model. 
We introduce \method, a causal transformer architecture with modality and temporal blockwise attention masking to simultaneously process multiple social cues across multiple participants and their temporal interactions. 
We train and evaluate \method\ on the Human-Human Commensality Dataset (HHCD), and demonstrate that using multiple modalities improves bite timing and speaking status prediction. 
Source code: \href{https://github.com/AbrarAnwar/masked-social-signals/}{https://github.com/AbrarAnwar/masked-social-signals/}.

\end{abstract}

\section{Introduction}\label{sec:introduction}
Human behavior in social interactions is shaped by a continuous dance of multimodal signals~\cite{kendon1970movement, lafrance1979nonverbal}.
There are dozens of social signals such as gesture, head orientation, gaze, and speech, that humans send during interactions.
In shared social settings like dining, where individuals engage in simultaneous verbal and non-verbal communication, these signals do not exist in isolation. 
Instead, each person’s behavior is intertwined with the actions of others.
A person’s gaze may shift in response to a conversational partner’s body language, or they may pause mid-sentence as another prepares to take a bite of food. 
Research on learning from groups~\cite{alameda2015salsa, joo2019towards} typically focuses on learning individual features such as facial features~\cite{ng2022learning}, head pose~\cite{zhou2022responsive}, body pose~\cite{ng2020you2me}, hands pose~\cite{ng2021body2hands}, end-of-turn prediction~\cite{lee2024online}, or bite timing~\cite{ondras2022human} from other interactants.
Understanding such social signals is important for affective computing~\cite{wang2022systematic, mathur2023towards}, social robotics~\cite{murray2022learning, vazquez2017towards}, navigation~\cite{dieberger2000social,holman2021watch}, or robot-assisted feeding~\cite{ondras2022human}.
However, when interacting in multi-party settings, many social signals occur simultaneously over time.
Thus, models of social behavior must be able to learn from diverse social signals while accounting for inherent temporal dependencies.

In this work, we propose a causal transformer model {\methodlong}, which leverages modality-specific and time-based blockwise attention masking so that a single model can predict and leverage multiple features of social signals. 
This architecture is capable of leveraging information from multiple modalities --- such as body pose, head orientation, gaze direction, speech, and others --- while also attending to some length of past history. 
By processing multimodal features over time, the model can learn the choreography of social signals and predict an individual's behaviors.

We leverage \method\ on the Human-Human Commensality (HHCD) dataset~\cite{ondras2022human} of triadic dining. 
The dataset was constructed for predicting when someone takes a bite of food, which is called bite timing, and contains various multimodal signals, such as speech, body pose, gaze, food interactions, and more.
In this work, we demonstrate that \method\ is able to predict an individual’s behavior based on the multimodal cues of others in their group.
This task, particularly bite timing, is an important problem as an autonomous feeding system can help people with mobility limitations be fed autonomously with a robot in social settings~\cite{bhattacharjee2020more}.
We then show the importance of including multiple modalities to predict these social signals, along with ablations on the role of larger temporal contexts and temporal chunking.

\begin{figure*}[t]
    \centering
    \includegraphics[width=0.75\textwidth]{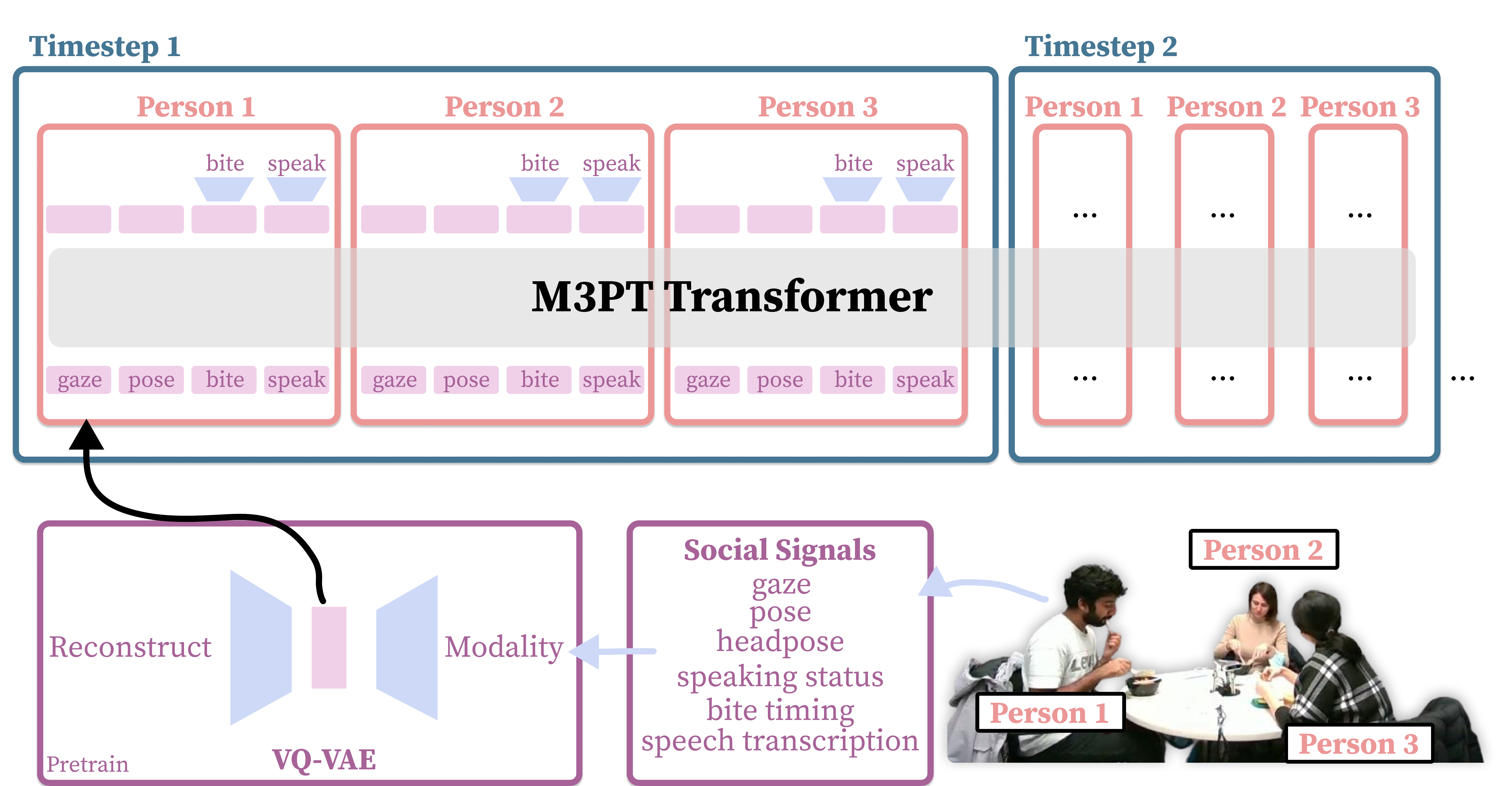}
    \caption{\textbf{Model Architecture.} 
        {\method} consists of a multi-party, multimodal causal transformer that attends to social signal features across people and timesteps.
        This transformer allows our model to reason about the interactants in multi-party settings over time.
        We encode these features with a VQ-VAE to tokenize continuous social signals for the transformer.
        We apply time-specific, person-specific, and modality-specific positional encodings along with a blockwise attenttion masking strategy, as shown in Fig.~\ref{fig:blockatten} to allow for the transformer to learn the relationship between these inputs.
        Then, we reconstruct the discrete social signals, speaking status and bite timing.
        }
    \label{fig:model}
\end{figure*}

\section{Multimodal Social Signal Prediction Task}\label{sec:problem_statement}


The goal of a social prediction task is to predict the social behaviors of a target person by using social cues from their interlocutors. 
Formally, at timestep $t$, an individual $i$ gives off signals $X_i(t)$.
$X_i(t)$ contains all the social signals a person gives off such as gaze, body gesture, and speech.
In a scenario with $n$ interactants, the social signals of person $i$ are dependent on those from other interactants $j, j \ne i$ from the current and past timesteps in addition to social signals of person $i$ from  previous timesteps.
Thus the goal of social signal prediction is to learn a function $\mathcal{F}$ that is able to predict the social signals of an individual $i$ at timestep $t$:
\begin{equation}
    X_i(t) = \mathcal{F}(X_i(0 : t-1), \{X_j(0 : t),  \forall j \ne i\}).
\end{equation}
Learned function $\mathcal{F}$ is meant to predict social signals of arbitrary participants.
We consider each timestep $t$ to be an interval of length $c$ as opposed to an absolute point of time.
For example, a social signal such as gaze at timestep $t$ is represented as a $c$-second gaze signal segment.
Then, each of these segments can be downsampled to any framerate.

We note that there is a distinction between social signal prediction and social signal forecasting, as the latter involves predicting signals in some time frame in the future, while prediction involves predicting concurrent signals.
In this work, we limit our scope to signal prediction, which includes conditioning on concurrent features from other interactants.

\begin{figure}[b]
    \centering
    \includegraphics[width=\linewidth]{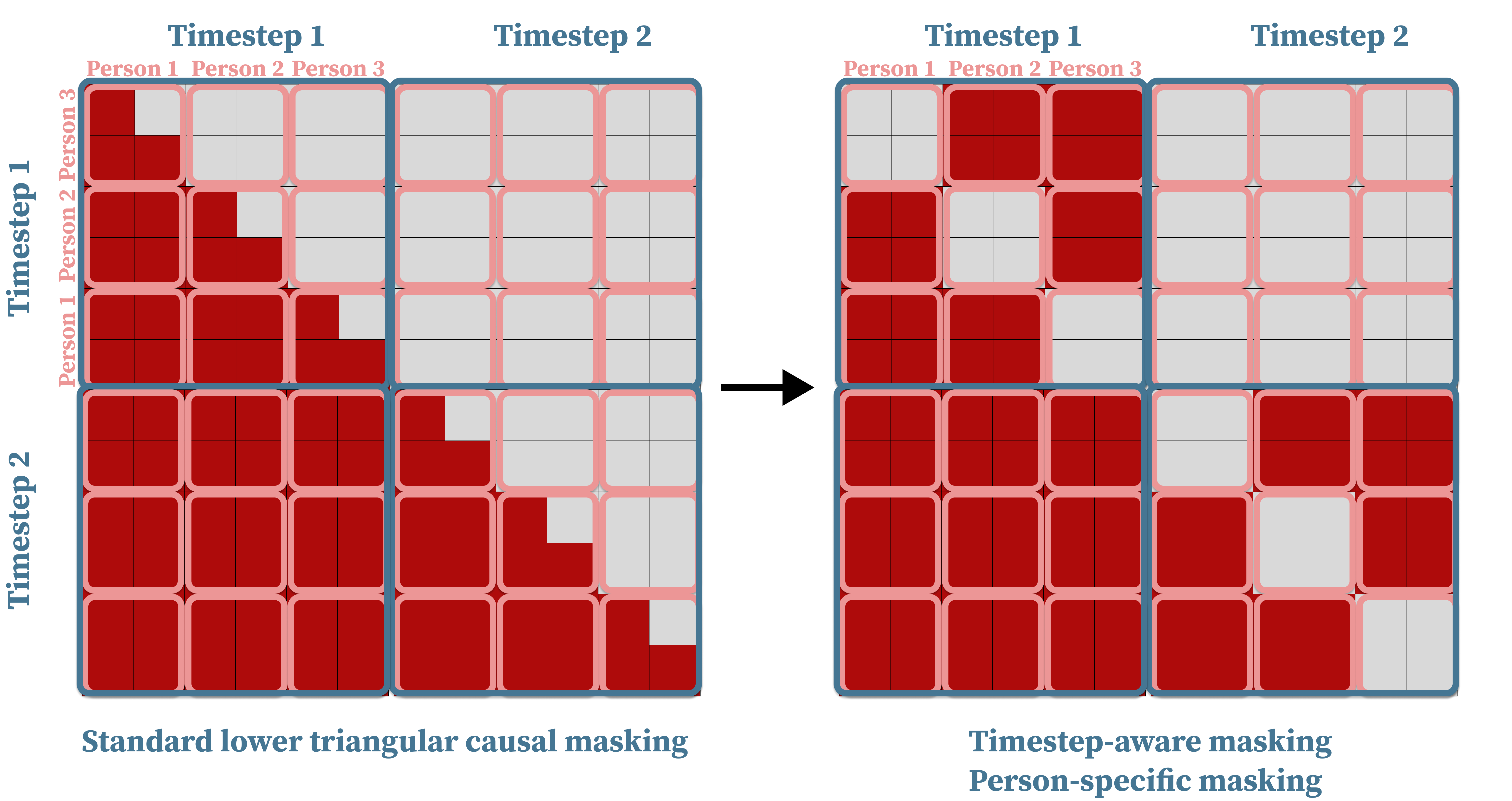}
    \caption{\textbf{Blockwise Attention Masking}:
    Conventional lower-triangular masking (left) would not be aware of the timestep and person-specific feature chunks present in multi-party, multimodal settings. 
    Our blockwise attention mask (right) masks each person's social signals when they must be predicted, allowing the model to focus solely on capturing interactions from others' social signals. 
    }
        
    \label{fig:blockatten}
\end{figure}

\section{\methodlong}\label{sec:model}
We introduce \methodlong, a model designed for multi-party social signal prediction. Unlike previous work that focuses on predicting a single social signal with a small time horizon or a single social signal, \method\ simultaneously considers  and predicts multiple social signals.

\method\ is designed to predict discrete social signals, but can handle a wide array of multimodal social signals, such as gaze, head pose, pose, and speech transcriptions as inputs.
We handle the diverse number of inputs by tokenizing each input modality from a person by pretraining a vector-quantized autoencoder for each signal.
Then, to learn the temporal relationship of these social signals across multiple interlocutors, we introduce a person-aware and modality-aware blockwise attention masking approach to attend over these tokens across modality, time, and interlocutor.


\subsection{Learning Modality-Specific Quantized Codebooks}

Because we are operating on $c$-second time segments, it is difficult to learn features for continuous features.
We tokenize continuous inputs with a Vector Quantized Variational Autoencoder (VQ-VAE) \cite{vqvae}. 
Signals like gaze, headpose, or pose can be represented as keypoints, and we use a 1D CNN-based encoder and decoder to learn quantized keypoint representations.
We first train these modality-specific VQ-VAEs on the HHCD training data, then freeze them during the training of the \method\ transformer.

To train the VQ-VAE, we take a $c$-second segment and encode each individual keyframe with a 1D-CNN. 
If the segment contains $m$ frames, we then have $m$ embeddings to represent the segment.
Each keyframe's embedding is then quantized by selecting from a codebook, which provides a discrete representation of the social signal.
Given these $m$ quantized embeddings, we aggregate them into a single embedding $z$ with a linear projection step.
During the decoding step, we up-project $z$ back into $m$ temporal embeddings, from which we use the codebook to re-select $m$ quantized embeddings.
We train the VQ-VAE with standard selection and commitment losses~\cite{vqvae}.
Since we leverage the quantization step in both the encoding step and the decoding step to learn $z$, we average the two selection losses and add a reconstruction loss. 
Thus, the total training objective becomes:
$\mathcal{L} = \log p(x|z_q(x)) + 0.5(\mathcal{L}^1_{\text{select}} + \mathcal{L}^2_{\text{select}}).$




\subsection{Transformer Architecture}

In a multi-party setting, each participant produces various social signals at each timestep.
We use a VQ-VAE to compute discretized tokens for each $c$-second segment in time. 
The encoded representations of all social signals from all individuals are proccessed by a causal (left-to-right) transformer designed to capture relationships between individuals and modalities over time.
For a modality $k$ for person $i$ over time $z^{ik}_0, \cdots, z_T^{ik}$, we add a cyclic positional encoding over time, an embedding representing the person, and a modality-specific embedding. 

We sequence the modalities over time per-person as shown in Fig.~\ref{fig:model}, where each timestep block contains blocks for the modalities of each individual. 
However, this blockwise structure is not applicable for the traditional lower triangular attention mask typically used in causal transformers. 
In particular, a lower triangular causal attention allows for a model to leverage past information to predict a given token; however, we want to predict the features of a person $i$ based on features from previous timesteps \textit{and} features from other participants in the current timestep.
Thus, we introduce multi-party, multimodal blockwise causal masking.

\begin{table}[t]
    \caption{\textbf{Bite Timing Prediction} performance across ablated input modalities. 
    Using all available signals leads to the best performance, and that dropping some signals causes substantial performance loss across all metrics, confirming the value of multimodality in predicting bite timing.
    }
    \centering
    \begin{tabular}{lcccc}
        \toprule
        Features & \textbf{F1} & \textbf{Precision} & \textbf{Recall} & \textbf{nMCC} \\ 
        \midrule
        All Features & 0.86{\scriptsize $\pm$ 0.06} & 0.81{\scriptsize $\pm$ 0.13} & 0.93{\scriptsize $\pm$ 0.04} & 0.92{\scriptsize $\pm$ 0.03} \\ \midrule
        No Gaze & 0.61{\scriptsize $\pm$ 0.35} & 0.60{\scriptsize $\pm$ 0.33} & 0.62{\scriptsize $\pm$ 0.37} & 0.78{\scriptsize $\pm$ 0.19} \\ 
        No Headpose & 0.83{\scriptsize $\pm$ 0.17} & 0.81{\scriptsize $\pm$ 0.26} & 0.91{\scriptsize $\pm$ 0.00} & 0.91{\scriptsize $\pm$ 0.08} \\ 
        No Pose & 0.89{\scriptsize $\pm$ 0.06} & 0.84{\scriptsize $\pm$ 0.13} & 0.97{\scriptsize $\pm$ 0.03} & 0.94{\scriptsize $\pm$ 0.03} \\ 
        No Word & 0.73{\scriptsize $\pm$ 0.30} & 0.73{\scriptsize $\pm$ 0.36} & 0.77{\scriptsize $\pm$ 0.19} & 0.84{\scriptsize $\pm$ 0.18} \\ 
        No Speaker & 0.24{\scriptsize $\pm$ 0.06} & 0.16{\scriptsize $\pm$ 0.04} & 0.46{\scriptsize $\pm$ 0.12} & 0.56{\scriptsize $\pm$ 0.02} \\ 
        \midrule
        Bite Only & 0.44{\scriptsize $\pm$ 0.35} & 0.39{\scriptsize $\pm$ 0.36} & 0.59{\scriptsize $\pm$ 0.29} & 0.69{\scriptsize $\pm$ 0.19} \\ 
        \bottomrule
    \end{tabular}
    
    \label{tab:bite_main}
\end{table}



\textbf{Blockwise Attention Masking.}
To enforce temporal structure and modality-specific interactions, we design a blockwise causal masking strategy. 
This mask consists of a lower triangular matrix that restricts attention to only past and present time steps, preserving the left-to-right structure of the model. 
Additionally, we modify the mask by creating small blocks along the diagonal, where signals from each individual can attend to both their own previous signals and those of others in the group as shown in Fig.~\ref{fig:blockatten}. 
This design ensures that the model can attend not only each person’s behavior but also to the potential influence of others' social signals on that behavior.

\textbf{Right-shifted Residual Connection.}
In bidirectional encoder architectures like BERT \cite{bert}, mask tokens prevents residual connections from leaking information to masked positions. 
Similarly, in unidirectional decoder architectures like GPT \cite{gpt}, predicting the next token inherently avoids data leakage through residual connections. 
However, in our case, our blockwise attention mechanism prevents the direct application of conventional residual connection strategies. 
We found that simply removing residual connections leads to poor model training due to small gradients.

To address this difficulty, we use a right-shifted residual connection. 
Instead of directly adding residual features to the hidden states, we right-shift all features by one segment before adding them. 
This shift ensures that each position receives residual information only from the preceding segment, which prevents leakage of the signal to be predicted during inference for evaluation.

\begin{table}[t]
    \caption{\textbf{Speaking Status Prediction}.
    We find that the best results for speaking status prediction are achieved when all features are included.
    We find that word and bite features are most predictive of predicting speaking status.
    }
    \centering
    \begin{tabular}{lcccc}
        \toprule
        Features & \textbf{F1} & \textbf{Precision} & \textbf{Recall} & \textbf{nMCC} \\ 
        \midrule
        All Features & 0.91{\scriptsize $\pm$ 0.03} & 0.86{\scriptsize $\pm$ 0.09} & 0.97{\scriptsize $\pm$ 0.04} & 0.94{\scriptsize $\pm$ 0.02} \\ \midrule
        No Gaze & 0.83{\scriptsize $\pm$ 0.05} & 0.73{\scriptsize $\pm$ 0.08} & 0.97{\scriptsize $\pm$ 0.02} & 0.89{\scriptsize $\pm$ 0.03} \\ 
        No Headpose & 0.85{\scriptsize $\pm$ 0.03} & 0.75{\scriptsize $\pm$ 0.05} & 1.00{\scriptsize $\pm$ 0.00} & 0.90{\scriptsize $\pm$ 0.02} \\ 
        No Pose & 0.89{\scriptsize $\pm$ 0.01} & 0.81{\scriptsize $\pm$ 0.03} & 0.99{\scriptsize $\pm$ 0.00} & 0.93{\scriptsize $\pm$ 0.01} \\ 
        No Word & 0.75{\scriptsize $\pm$ 0.18} & 0.66{\scriptsize $\pm$ 0.19} & 0.87{\scriptsize $\pm$ 0.17} & 0.83{\scriptsize $\pm$ 0.13} \\ 
        No Bite & 0.77{\scriptsize $\pm$ 0.16} & 0.67{\scriptsize $\pm$ 0.16} & 0.90{\scriptsize $\pm$ 0.12} & 0.84{\scriptsize $\pm$ 0.11} \\ \midrule
        Speaker Only & 0.89{\scriptsize $\pm$ 0.15} & 0.88{\scriptsize $\pm$ 0.16} & 0.90{\scriptsize $\pm$ 0.13} & 0.92{\scriptsize $\pm$ 0.10} \\ 
        \bottomrule
    \end{tabular}
    
    \label{tab:speaking_main}
\end{table}


\section{Evaluation on HHCD Dataset}\label{sec:evaluation}

To evaluate \method, we use the Human-Human Commensality Dataset (HHCD), which contains triadic interactions among three participants without mobility limitations during shared meals. 
The purpose of the dataset is to predict bite timing events in social settings, which can be used to build socially-aware robot-assisted feeding systems for individuals who do have mobility limitations.
The dataset contains multimodal social data like gaze, body pose, speech, as well as annotations for bite events, drink interactions, and utensil usage.
In this work, we repurpose this dataset towards our task of multi-party, multimodal social signal prediction.
We process gaze, headpose, body pose, transcripted speech, speaking status, and bite timing as inputs to \method, and we focus on improving the prediction of the two binary social signals of an individual given their past social signals and features of the other two participants: speaking status and bite timing.

HHCD contains 30 unique sessions of triadic social dining. 
For each session, we sample 36-second sequences with an 18-second rolling window. 
We then split this 36-second sequence into 12 three-second segments. 

{\method} performs binary classification tasks for both speaking status and biting time over each segment. 
Since each segment is 3-second long, we classify a segment as ``speaking" if more than 30\% of the frames indicate speaking for a specific user. This is to account for noise in the speaker estimations.
A frame is labeled as ``biting" if at least one frame within the segment has a bite. 




\textbf{Training objective.}
Our training objective is to minimize cross entropy losses for bite timing and speaking status prediction.
In HHCD, speaking status and biting time are highly imbalanced, as most of the time individuals are neither speaking nor biting. To address this imbalance, we apply inverse class frequency weighting to the loss functions. 

\textbf{Evaluation metrics.}
HHCD contains 30 triadic sessions; we treat each session as a fold and train on 29 of them and test on 1. 
We repeat this across 3-folds, training and testing on different sets of sessions.
We evaluate our models on classification accuracy, F1 score, precision, recall, and a normalized Matthews correlation coefficient (nMCC).
Unlike F1 score, nMCC summarizes the full confusion matrix of true/false positives and negatives, which past work~\cite{ondras2022human} found informative for bite timing.

\begin{table}[t]
     \caption{\textbf{Does \method\ learn from longer temporal context?}. We maintain the segment size of each token to be 3 seconds, and have $n$ segments $\times$ $3$ seconds-per-segment. Functionally, this ablation studies how \method\ makes predictions with longer contexts. We find that as the total context time considered increases, the F1 score begins to fall.}
    \label{tab:ablation_study}
    \centering
    \begin{tabular}{c@{}ccccc}
        \toprule
 & Task & \textbf{F1 Score} & \textbf{Precision} & \textbf{Recall} & \textbf{nMCC} \\ 
\midrule
        \multirow{2}{*}{2$\times$3s} & S & 0.99{\scriptsize $\pm$ 0.00} & 0.99{\scriptsize $\pm$ 0.00} & 1.00{\scriptsize $\pm$ 0.00} & 0.99{\scriptsize $\pm$ 0.00} \\ 
         & B & 0.99{\scriptsize $\pm$ 0.00} & 0.99{\scriptsize $\pm$ 0.01} & 1.00{\scriptsize $\pm$ 0.00} & 0.99{\scriptsize $\pm$ 0.00} \\ \midrule
        \multirow{2}{*}{3$\times$3s} & S & 0.99{\scriptsize $\pm$ 0.00} & 0.99{\scriptsize $\pm$ 0.00} & 1.00{\scriptsize $\pm$ 0.00} & 0.99{\scriptsize $\pm$ 0.00} \\ 
        & B & 0.97{\scriptsize $\pm$ 0.02} & 0.94{\scriptsize $\pm$ 0.04} & 1.00{\scriptsize $\pm$ 0.00} & 0.98{\scriptsize $\pm$ 0.01} \\ \midrule
        \multirow{2}{*}{6$\times$3s} & S & 1.00{\scriptsize $\pm$ 0.00} & 1.00{\scriptsize $\pm$ 0.00} & 1.00{\scriptsize $\pm$ 0.00} & 1.00{\scriptsize $\pm$ 0.00} \\ 
        & B & 1.00{\scriptsize $\pm$ 0.00} & 1.00{\scriptsize $\pm$ 0.00} & 1.00{\scriptsize $\pm$ 0.00} & 1.00{\scriptsize $\pm$ 0.00} \\ \midrule
         \multirow{2}{*}{12$\times$3s} & S & 0.91{\scriptsize $\pm$ 0.03} & 0.86{\scriptsize $\pm$ 0.09} & 0.97{\scriptsize $\pm$ 0.04} & 0.94{\scriptsize $\pm$ 0.02} \\ 
        & B & 0.86{\scriptsize $\pm$ 0.06} & 0.81{\scriptsize $\pm$ 0.13} & 0.93{\scriptsize $\pm$ 0.04} & 0.92{\scriptsize $\pm$ 0.03} \\ 
        \bottomrule
    \end{tabular}
    
\end{table}
\section{Results}\label{sec:results}

We present the test set performance of \method.
We report the average performance and standard deviation of \method\ and various ablations over three folds of the HHCD data.

\textbf{Multimodality improves social signal prediction performance.}
As shown in Tables~\ref{tab:bite_main} and~\ref{tab:speaking_main}, using all the modalities improves performance, both for speaking status and bite timing. 
For bite timing prediction, we find large distinctions across features. 
We find that all metrics, especially F1 and nMCC are reduced when gaze, words, or speaking status are removed. 
This social signals are empirically the most informative for predicting bite timing.

We find similar results indicating that multiple modalities improves speaking status prediction.
We find that using no other signals except for speaking status, the \method\ produces nearly random predictions, but using all modalities improves the performance the most.
The largest performance degradation occurs when bite timing is removed, indicating that it is an important feature for speaking status prediction.
Other features have slight performance losses, however, they are not as strong of a drop-off compared to bite timing.

\textbf{Temporal context on bite timing.}
Table~\ref{tab:ablation_study} presents an ablation that explores how varying the total time length, while maintaining a constant segment size, affects model performance for speaking status (S) and bite timing (B) tasks. 
The experiment is conducted using a smaller model to avoid overparameterization.
The results show that reducing the temporal context from 36 seconds to lower values actually increases performance, possibly due to task difficulty increasing as data gets noisier.
However, we believe that this 36 second length is practical for situations like predicting bite timing, where a system has to run continuously.

\textbf{Large segment lengths lead to mode collapse.}
Table~\ref{tab:ablation_segmenting} shows that, as the segment length increases, both speaking status and bite timing prediction accuracy drastically falls across all metrics. 
This finding is consistent with mode collapse in the tokenization step: the VQ-VAE step is representing more of the temporal information as opposed to the transformer.
Our use of 12 segments with a length of 3 seconds strikes a balance between having the transformer represent the temporal and multi-party information and having the VQ encoders represent only the modality information.

\begin{table}[t]
     \caption{
    \textbf{Impact of Segment Length on Performance.} We hold the total time to be at a constant 36 seconds, but modify the length of a segment to $k$ seconds and thus the number of segments $n$ to have $n\times k$s. We find that longer segment lengths leads to a collapse in prediction performance in speaking status (S) and bite timing (B) to majority class.
    }
    \centering
    \begin{tabular}{c@{}ccccc}
        \toprule
            & Task & \textbf{F1 Score} & \textbf{Precision} & \textbf{Recall} & \textbf{nMCC} \\ 
        \midrule
        \multirow{2}{*}{2$\times$18s} & S & 0.00{\scriptsize $\pm$ 0.00} & 0.00{\scriptsize $\pm$ 0.00} & 0.00{\scriptsize $\pm$ 0.00} & 0.00{\scriptsize $\pm$ 0.00} \\ 
         & B & 0.64{\scriptsize $\pm$ 0.04} & 0.47{\scriptsize $\pm$ 0.05} & 1.00{\scriptsize $\pm$ 0.00} & 0.64{\scriptsize $\pm$ 0.00} \\ \midrule
        \multirow{2}{*}{4$\times$9s} & S & 0.12{\scriptsize $\pm$ 0.18} & 0.08{\scriptsize $\pm$ 0.12} & 0.25{\scriptsize $\pm$ 0.35} & 0.51{\scriptsize $\pm$ 0.00} \\ 
        & B & 0.43{\scriptsize $\pm$ 0.08} & 0.27{\scriptsize $\pm$ 0.06} & 1.00{\scriptsize $\pm$ 0.00} & 0.52{\scriptsize $\pm$ 0.05} \\ \midrule
        \multirow{2}{*}{6$\times$6s} & S & 0.23{\scriptsize $\pm$ 0.17} & 0.15{\scriptsize $\pm$ 0.11} & 0.50{\scriptsize $\pm$ 0.37} & 0.50{\scriptsize $\pm$ 0.01} \\ 
        & B & 0.32{\scriptsize $\pm$ 0.08} & 0.19{\scriptsize $\pm$ 0.06} & 0.99{\scriptsize $\pm$ 0.00} & 0.52{\scriptsize $\pm$ 0.03} \\ \midrule
        \multirow{2}{*}{12$\times$3s} & S & 0.91{\scriptsize $\pm$ 0.03} & 0.86{\scriptsize $\pm$ 0.09} & 0.97{\scriptsize $\pm$ 0.04} & 0.94{\scriptsize $\pm$ 0.02} \\ 
        & B & 0.86{\scriptsize $\pm$ 0.06} & 0.81{\scriptsize $\pm$ 0.13} & 0.93{\scriptsize $\pm$ 0.04} & 0.92{\scriptsize $\pm$ 0.03} \\ 
        \bottomrule
    \end{tabular}
   
    \label{tab:ablation_segmenting}
\end{table}

\section{Conclusion and Limitations}\label{sec:future}

In this work, we presented \method, a causal transformer able to predict an individual's bite timing and speaking status in multi-party dining based on temporal- and person-aware social signals.
We investigated our design decision for 3-second temporal segments, and found that the larger segments cause the transformer to learn less of the temporal structure.
We also presented results on the impact of increased temporal context for predicting btie timing and speaking status.
\method\ is a first step in building models for multi-party, multimodal social signal processing.

If we want to build a robot that can emulate various social signals, we need to predict continuous social signals in addition to discrete signals.
In preliminary experiments, we found that predicting continuous signals like body pose directly was not feasible. 
The reconstructions were often poor as transformer models are often better at predicting discretized inputs.
When trained to predict discrete pose tokens constructed by the VQ-VAE, the \method\ performance was still lacking, though it is unclear whether the behavior of one's interlocutorrs is a sufficient signal to predict what body pose one will take next. 
We plan to explore predicting continuous social signals in future work such as predicting body pose, gaze, etc.

\section{Ethical Impact Statement.}

\method\ was designed to utilize social signals in multi-party settings.
Although this work had used pre-collected data from others' work, the implication of a model such as \method\ has potential uses in tracking users and potentially manipulating social scenarios.
Additionally, if such a method can process social signals from others who do not know they are expressing such information, consent of the utilization of such data becomes important. 
Otherwise, such a social signal tracking system could be used for unknowingly manipulating non-consenting participants.
Also, the original application of bite timing is to use this method for helping people with mobility limitations.
Though this work does not explicitly consider this setting, we must be careful in introducing bias in training such models, as they would be used in sensitive situations with people with mobility limitations.
For example, body pose prediction may introduce bias for systems that run alongside people with mobility limitations, and these kinds of considerations need to be made carefully.

To mitigate such risks, the utilization of \method\ can require explicit consenting procedures or only track users who have previously consented to be observed or can rely only on anonymized data such as body pose that were processed through anonymized pipelines.
However, we believe that the benefits of such a system are still useful.
For bite timing prediction, past work~\cite{bhattacharjee2020more} has shown that people with mobility limitations who use robot-assisted feeding systems prefer such autonomous systems as they are able to eat food with their friends and families.

\nocite{*}

\bibliographystyle{ieee}
\bibliography{egbib}

\end{document}